# Efficient Machine Learning for Big Data: A Review

O. Y. Al-Jarrah [a], P. D. Yoo [b], S Muhaidat [c], G. K. Karagiannidis [a, d], and K. Taha [a]

[a] *Khalifa University, Abu-Dhabi, UAE*

[b] *Data Science Institute, Bournemouth University, UK*

[c] *University of Surrey, Guildford, UK*

[d] *Aristotle University of Thessaloniki, Thessaloniki, Greece*

**Abstract**

With the emerging technologies and all associated devices, it is predicted that massive amount of data will be created in the next few years – in fact, as much as 90% of current data were created in the last couple of years – a trend that will continue for the foreseeable future. Sustainable computing studies the process by which computer engineer/scientist designs computers and associated subsystems efficiently and effectively with minimal impact on the environment. However, current intelligent machine-learning systems are performance driven – the focus is on the predictive/classification accuracy, based on known properties learned from the training samples. For instance, most machine-learning-based nonparametric models are known to require high computational cost in order to find the global optima. With the learning task in a large dataset, the number of hidden nodes within the network will therefore increase significantly, which eventually leads to an exponential rise in computational complexity. This paper thus reviews the theoretical and experimental data-modeling literature, in large-scale data-intensive fields, relating to: (1) model efficiency, including computational requirements in learning, and data-intensive areas' structure and design, and introduces (2) new algorithmic approaches with the least memory requirements and processing to minimize computational cost, while maintaining/improving its predictive/classification accuracy and stability.

## 1. Introduction

Today, it's no surprise that reducing energy costs is one of the top priorities for many energy-related businesses. The global information and communications technology (ICT) industry that pumps out around 830 Mt carbon dioxide ($CO_2$) emission accounts for approximately 2 percent of the global $CO_2$ emissions [1]. ICT giants are constantly installing more servers so as to expand their capacity. The number of server computers in data centers has increased sixfold to 30 million in the last decade, and each server draws far more electricity than its earlier models [2]. The aggregate electricity use for servers had doubled between the years 2000 and 2005 period, most of which came from businesses installing large numbers of new servers [3]. This increase in energy consumption consequently results in higher carbon dioxide emissions, and hence causing an impact on the environment. Furthermore, most of these businesses, especially in an uncertain economic climate are placed under the pressure to reduce their energy expenditure in order to remain competitive in the market [4].

With the emerging of new technologies and all associated devices, it is predicted that there will be as much data created as was created in the entire history of planet Earth [5]. Given the unprecedented amount of data that will be produced, collected and stored in the coming years, one of the technology industry's great challenges is how to benefit from it. During the past decade, mathematical intelligent machine-learning systems have been widely adopted in a number of massive and complex data-intensive fields such as astronomy, biology, climatology, medicine, finance and economy. However, current intelligent machine-learning-based systems are not inherently efficient or scalable enough to deal with large volume of data. For example, for many years, it is known that most non-parametric and model-free approaches require high computational cost to find the global optima. With high-dimensional data, their good data fitting capacity not only makes them more susceptible to the generalization problem but leads to an exponential rise in computational complexity. Designing more accurate machine-learning systems so as to satisfy the market needs will hence lead to a higher likelihood of energy waste due to the increased computational cost.

Nowadays, there is a greater need to develop efficient intelligent models to cope with future demands that are in line with similar energy-related initiatives. Such energy-efficient-oriented data modeling is important for a number of data-intensive areas, as they affect many related industries. Designers should focus on maximum performance and minimum energy use so as to break away from the traditional' performance vs. energy-use' tradeoff, and increase the number and diversity of options available for energy-efficient modeling. However, despite the fact that there is a demand for such efficient and sustainable data modeling methods for large and complex data-intensive fields, to our best knowledge, only a few of these literatures have been proposed in the field [6][7].

This paper provides a comprehensive review of state-of-the-art sustainable/energy-efficient machine-learning literatures, including theoretical, empirical and experimental studies pertaining to the various needs and recommendations. Our objective is to introduce a new perspective for engineers, scientists, and researchers in the computer science, and green ICT domain, as well as to provide its roadmap for future research endeavors.

This paper is organized as follows. Section 2 introduces the different large-scale data-intensive areas and discusses their structure and nature, including the relation between data models and their characteristics. Section 3 discusses the issues in current intelligent data modeling for sustainability and gives recommendations. Section 4 concludes the paper.

## 2. Big data challenge

e-Science areas are typically data-intensive in that the quality of their results improves with both quantity and quality of data available. However, current intelligent machine-learning systems are not inherently efficient enough which ends up, in many cases, a growing fraction of this quantity data unexplored and underexploited. It is no small problem when existing methods fail to capture such data immensity. When old concepts fail to keep up with change, traditions and past experience become inadequate guide for what to do next. Effective understanding and the use of this new wealth of raw information pose a great challenge to today's green engineers/researchers. It should be noted that the scope of the review is limited to the analytical aspects of science areas using immense datasets, and the methods for reducing computational complexity in distributed or grid-computing environment is excluded.

### 2.1. Geo, climate and environment

There are many recent examples that can illustrate the tremendous growth in scientific data generation in the literature. It is estimated that there are thousands of wireless sensors currently in place, which generates about a gigabyte of data per sensor per day [8]. Such sensors measure and record sensory information about the natural environment at a joint spatial and temporal dimensions that has never previously been possible. This environmental information is gathered by sensors via its sensing devices that are attached to small, low-power computer systems with digital radio communications. The sensor nodes self-organize itself into a network to deliver, and perhaps process the collected data to a base station, where it can be made available to the users through the Internet. These sensors generate several petabytes of data per year and decisions need to be taken in real time as to how much data to analyze, how much to transmit for further analysis.

Besides the environmentalists, a similar challenge facing the climatologists, meteorologists, and geologists today is also making sense of the vast and continually increasing amount of data generated by the earth observation satellites, radars, and high-throughput sensor networks. The World Data Centre for Climate (WDCC) is the world-largest climate data repository, and is also known to have the largest database in the world [9]. The WDCC archives 340 terabytes of earth system model data and related observations, and 220 terabytes of data readily accessible on the web including information on climate research and anticipated climatic trends, as well as 110 terabytes (or 24,500 DVD's) worth of climate simulation data. The WDCC data is accessible by a standard web-interface (http://cera.wdc-climate.de). These data are increasingly available in many different formats and have to be incorporated correctly into the various climate change models. Timely and accurate interpretation of these data can provide advance warnings in times of severe weather changes, hence enabling corresponding action to be taken promptly so as to minimize its resulting catastrophic damage.

### 2.2. Bio, medicine, and health

Biological data has been produced at a phenomenal rate due to the international research effort called the Human Genome Project. It is estimated that the human genome DNA contains around 3.2 billion base (3.2 gigabase) pairs distributed among twenty-three chromosomes, which is translated to about a gigabyte of information [10]. However, when we add the gene sequence data (data on the 100,000 or so translated proteins and the 32,000,000 amino acids), the relevant data volume can easily expand to an order of about 200 gigabyte [11]. Now, by including also the X-ray/NMR spectroscopy structure determination of these proteins, the data volume will increase dramatically to several petabytes, and that is assuming only one structure per protein.

As of December 2014, the GenBank repository of nucleic acid sequences contained above 178 million entries [12] and the SWISS-PROT database (inc. both UniProtKB/Swiss-Prot, UniProtKB/TrEMBL) of protein sequences contained about 18 million entries [13][14]. On average, these databases are doubling in size in every 15 months. This is further compounded by data generated from the myriad of related projects that study gene expression, that determines the protein structures encoded by the genes, and that details how these proteins interact with one another. From that, we can begin to imagine the enormous amount and variety of information that is being produced every month.

Over the past decade, the health sector has also evolved significantly, from paper-based systems to largely paperless electronic systems. Many countries' public health systems are now providing electronic patient records with advanced medical imaging media. In fact, this has already been implemented by more than 200 American hospitals, and the days of squinting to decipher a doctor's untidy scrawl on a handwritten prescription will soon be a thing of the past in Canada and many other countries too [15].

InSiteOne is one of the leading service providers in offering data archiving, storage, and disaster-recovery solutions to the healthcare industry. Its U.S. InSiteOne's archives include almost 4 billion medical images and 60 million clinical studies, in a coverage area of about 800 clinical sites [16]. The combined annual total of its radiological images exceeds 420 million and this number is still increasing at an approximate rate of about 12% per year. There are about 35,500 radiologists currently practicing in the U.S [17]. Each image will typically constitute several megabytes of digital data and is required to be archived for a minimum of five years. ESG (Enterprise Storage Group) forecasts medical image data in North America will grow to more than 35 percent per year and will reach nearly 2.6 million terabytes by 2014 [18]. It is also worthwhile to note that for the digital health data, its integrity and security issues are of critical importance in the field. For instance, for the former, data compression techniques may not be used, in many cases, as they may distort the data; and for the latter, the confidentiality of patient data is clearly cardinal in order to foster public confidence in such technologies.

*2.3. Stars, galaxies, and the universe*

The digital data volume from the stars, galaxies and universe has multiplied over the past decade due to the rapid development of new technologies such as new satellites, telescopes and other observatory instruments. Recently, the Visible and Infrared Survey Telescope for Astronomy (VISTA) [19] and the Dark Energy Survey (DES) [20] – the largest universe survey projects initiated by two different consortiums of universities, from the U.K., and from the U.S., are expected to yield databases of 20–30 terabytes in size in the next decade.

According to DES, its observatory field is so large that a single image will record data from an area of the sky 20 times the size of the moon as seen from the earth [20]. The survey will image 5000 degrees of the U.S. southern sky and will take about five years to complete. As for VISTA, its performance requirements were so challenging that it peaks at 55 megabytes/second data rate with a maximum of 1.4 terabytes of data per night [19]. But, these are now fairly commonplace. Many other astro-scientific databases, such as the Sloan Digital Sky Survey (SDSS) are already terabytes in size [21] and the Panoramic Survey Telescope-Rapid Response System (Pan-STARRS) is expected to produce a science database of more than 100 terabytes in size for the next five years [22]. Likewise, the Large Synoptic Survey Telescope (LSST) is producing 30 terabytes of data per night, yielding a total database of about 150 petabytes [23]. As the data produced by the new telescopes are expected to come to the Internet, this picture will change radically.

Many believe that the massive data volume and the ever increasing computing power will dramatically change the way in how conventional science and technology are conducted. We believe that this surge in data will open up and challenge further research in each field, hence, instigating the search for new approaches. Likewise, such challenge needs to be addressed in the area of intelligent information science as well.

## 3. Sustainable data modeling and efficient learning

With consideration of the large influx of data, it is definitely necessary to improve the way in how conventional computational/analytic data models are designed and developed. Sustainable data modeling can be defined as a form of data modeling technology, aimed to make sense of the large amount of data associated in its own field, by discovering patterns and correlations in an effective and efficient way. Sustainable data modeling specifically focuses on 1) maximum learning accuracy with minimum computational cost, and 2) rapid and efficient processing of large volumes of data. Sustainable data modeling seems to be ideal because of its ease in which large quantities of data are handled efficiently as well as its associated cost reduction observed in many cases. In a wider perspective, it entails a data-modeling revolution in e-sciences. In fact, these newly designed sustainable data models will effectively cope with the above data issues and, as a result, bring about benefits to the various e-science areas. Some of the excellent examples are well discussed in Patnaik *et al.*, Sundaravaradan *et al.*, and Marwah's article [24–27]. Hence, in this section, we will give a few recommendations to green engineers/researchers on a few key mechanics of the sustainable data modeling.

*3.1. Ensemble models*

One of the key success elements of sustainable data modeling is to maintain or improve its performance while significantly reducing its computational cost. Recent data-modeling research has shown that ensemble methods have gained much popularity as they often perform better than individual models [28][29]. Ensemble method uses multiple models to obtain better performance than those that could be obtained from any of the constituent models [29][30]. However, it can result in significant increase in computational cost. If the model deals with large-scale data, model complexity and computational requirements will grow exponentially. An example of such ensemble model is the Bayes classifier [31]. In Bayes classifier, each hypothesis is given a vote proportional to the likelihood that the training dataset would be sampled from a system if that hypothesis was true. To facilitate the training data of finite size, the vote of each hypothesis is also multiplied by the prior probability of that hypothesis. The Bayes classifier is expressed as follows:

$$y = \arg\max_{c_j \in C} \sum_{h_i \in H} P(c_j \mid h_i) P(T \mid h_i) P(h_i),$$

where *y* is the predicted class, *C* is the set of all possible classes, *H* is the hypothesis space, *P* refers to a probability, and *T* is the training data. As an ensemble, the Bayes classifier represents a hypothesis that is not necessarily in *H*. The hypothesis represented by the Bayes classifier, however, is the optimal hypothesis in ensemble space (the space of all possible ensembles consisting only of hypotheses in *H*).

Considering the problem of numerical weather prediction, ensemble predictions are now commonly made at most of the major operational weather prediction facilities worldwide [32], including the National Centers for Environmental Prediction, U.S., the European Centre for Medium-Range

Weather Forecasts (ECMWF), the United Kingdom Met Office, Metro France, Environment Canada, the Japanese Meteorological Agency, the Bureau of Meteorology, Australia, the China Meteorological Administration, the Korea Meteorological Administration, and CPTEC, Brazil.

*3.2. Model complexity problem*

Bayes estimation techniques have been well-adopted in general intelligent data modeling because they provide a fundamental formalism for combining all the information available, with regards to the parameters to be estimated, with optimized time complexity [33].

One of the most serious problems in Bayes nonparametric learning models is its high-algorithmic complexity and extensive memory requirements, especially for the necessary quadratic programming in large-scale tasks. As a nonparametric Bayes classifier extracts worst-case example $x$ and uses statistical analysis to build a classifying model, any learning algorithm that examines every attribute values of every training example must have at least the same or worse complexity [33].

Many applications of machine learning deal with problems where both the number of features $i$ as well as the number of examples $x_i$ is large. Linear Support Vector Machines are among the most prominent machine-learning techniques for such high-dimensional and sparse data. In this article, we use two machine-learning models as examples to be semiparameterized. In other words, the two models are to be modified to be more efficient and fast computationally. The time complexity of the Bayes and SVMs are well discussed in Elkan's and Joachims' article respectively [34][35].

*3.3. Local learning strategy*

Yoo *et al*. have proposed two different support-vector-based efficient ensemble models that have shown to reduce its computational cost while maintaining its performance [36]. Their novel learning technique has proven to be successful by other similar studies [7]. With a nonparametric model, a unique model must be constructed for each test set, which will significantly increase its computational complexity and cost.

To reduce the computational cost, they have thus proposed to partition the training samples into clusters, with that, build a separate local model for each cluster – this method is called local learning. A number of recent works have demonstrated that such a local learning strategy is far superior to that of the global learning strategy, especially on data sets that are not evenly distributed [37–40]. If a local-learning method is adopted in the decision function of a nonparametric classifier (i.e., the general regression network), it will allow for the classifier to be semiparameterized. Its semiparametric approximation can be expressed as follows:

$$Z_i \exp\frac{-(x-c_i)^T(x-c_i)}{2\sigma^2} \approx \sum_{j=1}^{Z_i} \exp\frac{-(x-x_j)^T(x-x_j)}{2\sigma^2},$$

where $x_i$ is a training vector for class $i$ in the input space, $\sigma$ is a single learning or smoothing parameter chosen during the network training, and $Z_i$ is a number of input training vectors $x_i$ associated with its center $c_i$. In nonparametric classification, many different types of radial basis functions can be chosen in place of the Gaussian function. The radial basis function, used in many cases, is actually a spherical kernel function, which is specifically used for nonparametric function estimation. If the number of training samples approaches infinity, the nonparametric function estimation hence becomes no longer dependent on the parameters of the radial basis function, however, for finite training samples, we can always observe some forms of dependency on the radial basis function parameters.

The local learning strategy provides more dependence on the radial basis function parameters than that of a nonparametric model because the local learning model is a semiparametric approximation of a nonparametric/global learning model. In other words, in semiparametric modeling, model assumptions gets stronger than those of nonparametric models, but are less restrictive than those of parametric model. In particular, this approximation avoids the practical disadvantages of nonparametric methods at the expense of increased risk of specification errors. Semiparametric models that are based on local learning help not only in reducing the model complexity but also in finding the optimal tradeoff between the parametric and nonparametric models – so as to achieve both low model bias and variance [41]. In short, it can therefore take on the inherent advantage of both the models while reducing its computational requirements effectively.

*3.4. Semiparametric approximation*

The above examples can be seen as a spherical function mixture model with data-directed center vector allocation. That is because the relative widths of the spherical functions at each center are directly proportional to the relative number of training vectors associated with each center. Many different types of computational local models, and the diverse selection method of the *yi* and the grouping of the associated input vectors in each class *i* can be used for the global model semiparametric approximation.

The local learning strategy provides a reasonable approximation since $x_i$ are sufficiently close in the input vector space. In that case, they can be adequately represented by a single center vector $c_i$ in that local space. In the case of Support Vector Regression (SVR), the $c_i$ vectors can be derived from either the *k*-means or the codebook theory. In SVR, where the two classes are not separable, they map the input space into a high-dimensional feature space (where the classes are linearly separable), using a nonlinear kernel function. The kernel function calculates the scalar product of the images of two examples in the feature space.

Given a *n*-dimensional input vector, $x_i=(x_1,x_2,...,x_n)$ with two labels, $y_i \in \{+1, -1\}$ where $i=1,2,...,N$, the hyperplane decision function of the binary SVR with kernel method is:

$$f(x) = \text{sgn}\left(\sum_{i=1}^{\ell} y_i a_i \langle \Phi(x), \Phi(x_i) \rangle + b\right) = \text{sgn}\left(\sum_{i=1}^{\ell} y_i a_i k(x, x_i) + b\right)$$

and the quadratic program is given as:

$$\text{maximize } W(a) = \sum_{i=1}^{\ell} a_i - \frac{1}{2} \sum_{i,j=1}^{\ell} a_i a_j y_i y_j k(x_i, x_j),$$

subject to $a_i \geq 0$, $i = 1,...,\ell$, and $\sum_{i=1}^{\ell} a_i y_i = 0$,

where $\ell$ is the number of training patterns, $a_i$ is the parameters of SVR, $K(.,.)$ is a spherical (nonparametric) kernel function, and $b$ is the bias term. In the above case, the local model can be constructed from $k$-means clustering. The objective function of the $k$-means clustering can be expressed as follows:

$$\min_{C,Z} \sum_{j=1}^{k} \sum_{i=1}^{n} Z_{i,j} \|X_i - C_j\|_2^2 + R \sum_{j=1}^{k} \left| \sum_{i=1}^{n} Z_{i,j} y_i \right|,$$

where $X_i$ is the $i^{th}$ row of the similarity matrix $\Sigma$, $C_j$ is a $1 \times m$ row vector representing the centroid of the $j^{th}$ cluster, $R$ is a non-negative scaling parameter, and $Z_{ij} \in \{0,1\}$ is an element of the cluster membership matrix, whose value is equal to one if the $i^{th}$ source vector belongs to the $j^{th}$ cluster, and zero if otherwise. The first term in the objective function corresponds to a cluster cohesion measure. The minimization of the above equation would ensure that the training vectors in the same cluster have highly correlated similarity vectors. The second term measures the skewness of class distribution in each cluster. The minimization of this term would ensure that each cluster contains a balanced number of positive and negative estimation vectors. The cluster centroid $C$ and cluster membership matrix $Z$ are estimated iteratively as follows:

- We fix the cluster centroids and use them to determine the cluster membership matrix.
- The revised cluster membership matrix is used to update the centroids. – repeated until the algorithm converges to a local minimum.

To compute the cluster membership matrix $Z$, we transform the original optimization problem, using $k$ slack variable $t_j$, into:

$$\min_{Z,t} \sum_{j=1}^{k} \sum_{i=1}^{n} Z_{i,j} \|X_i - C_j\|_2^2 + R \sum_{j=1}^{k} t_j,$$

**s.t.** $-t_j \leq \sum_{i=1}^{n} Z_{i,j} y_i \leq t_j,$

$t_j \geq 0, 0 \leq Z_{i,j} \leq 1,$

$\sum_{j=1}^{k} Z_{i,j} = 1,$

if the cluster membership matrix is obtained, the cluster centroid $C_j$ is updated based on the following:

$$Q_j(X_m) = C_j = \frac{\sum_{i=1}^{n} Z_{i,j} X_i}{\sum_{i=1}^{n} Z_{i,j}}, j = 1,2,...,N.$$

To construct a semiparametric model, we substituted $Q_i(X)$ for each training sample $x_i$ used in the SVR decision function. The new semiparametric model's approximation is therefore expressed as:

$$f(x) = \text{sgn}(w \cdot \phi(x) + b) = \text{sgn}\left( \sum_{i=1}^{\ell} y_i a_i k(x, c_i) + b \right),$$

and the quadratic program is given as:

$$\text{maximize } W(a) = \sum_{i=1}^{\ell} a_i - \frac{1}{2} \sum_{i,j=1}^{\ell} a_i a_j y_i y_j k(Q_i(x), Q_j(x)),$$

subject to $a_i \geq 0$, $i = 1,...,\ell$, and $\sum_{i=1}^{\ell} a_i y_i = 0.$

As mentioned, the local model can also be constructed from the principle of codebook [42]. In this case, its basic idea is to replace key values from an original multidimensional vector space with values from a discrete subspace of lower dimension. The lower-dimension vector requires less storage space and the data is thus compressed.

Consider a training sequence consisting of $M$ source vectors, $T=\{x_1, x_2, ..., x_m\}$. $M$ is assumed to be sufficiently large, such that all the statistical properties of the source are captured by the training sequence. We assume that the source vectors are $k$-dimensional, $X_m=(x_{m,1}, x_{m,2}, ..., x_{m,k})$, $m=1,2,...,M$. These vectors are compressed by choosing the nearest matching vectors, and form a codebook comprising of the entire set of codevectors. $N$ is the number of codevectors, $C=\{c_1,c_2,...,c_n\}$ and each codevector is $k$-dimensional, $c_n=(c_{n,1},c_{n,2},...,c_{n,k})$, $n=1,2,...,N$. The representative codevector is determined to be the closest in Euclidean distance from the source vector. The Euclidean distance is defined by:

$$d(x, c_i) = \sqrt{\sum_{j=1}^{k} (x_j - c_{ij})^2},$$

where $x_j$ is the $j^{th}$ component of the source vector, $c_{ij}$ is the $j^{th}$ component of the codevector $c_i$, $S_n$ is the nearest-neighboring region associated with codevector $c_n$, and the partitions of the whole region are denoted by $P=\{S_1,S_2,...,S_N\}$. If the source vector $X_m$ is in the region $S_n$, its approximation can be denoted by $Q(X_m)=c_n$, if $X_m \in S_n$. The Voronoi region is defined by:

$$V_i = \{x \in R^k : \|x - c_i\| \leq \|x - c_j\|, \text{ for all } j \neq i\},$$

the training vectors falling into a particular region are approximated by a red dot associated with that region (Fig. 1.).

To find the optimal $C$ and $P$, vector quantization uses a square-error distortion measure that specifies exactly how close the approximation is. The distortion measure is given as:

$$D_{ave} = \frac{1}{Mk} \sum_{m=1}^{M} \|X_m - Q(X_m)\|^2$$

If *C* and *P* are solution parameters to the minimization problem, then it must satisfy two conditions: (1) nearest-neighbor and (2) centroid. The nearest-neighbor condition indicates that the subregion $S_n$ should consist of all the vectors that are closer to $c_n$ than any of the other codevectors:

$$S_n = \{x : \|x - c_n\|^2 \leq \|x - c_{n'}\|^2, \forall n' = 1, 2, ..., N\},$$

finally, the centroid condition indicates that the codevector $c_n$ can be derived from the average of all the training vectors in its Voronoi Region $S_n$:

$$c_n = \frac{\sum_{Xm \in Sn} X_m}{\sum_{Xm \in Sn} 1}, n = 1, 2, ..., N.$$

As Elkan's discussed [34], the local learning techniques – use of $c_n$ vectors for building a local model – prove that any intelligent learning model that examines all the attribute values of every training example must have the same or worse complexity. In other words, such a local learning strategy is far more efficient than that of the global learning strategy, especially on a large volume of data problems [37–40].

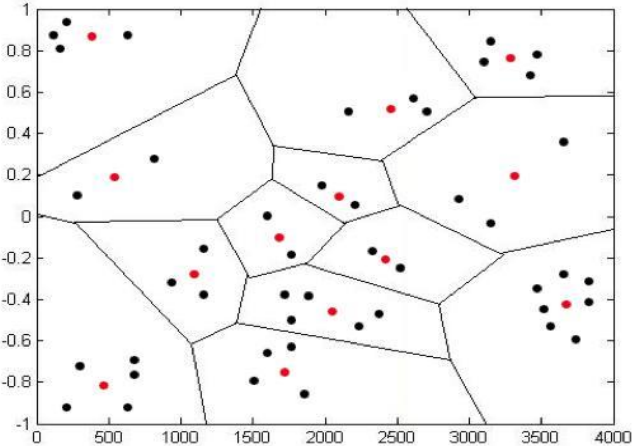

**Fig. 1.** Two-dimensional (2D) vector quantization

### 3.5. Deep learning

Shallow learning models (e.g., SVM, MLP, and GMM) have been widely used in the literature to solve simple or well-constrained problems. However, their limited modeling and representational power do not support their use in solving more complex problem, such as natural language problems. In 2006, the so-called deep learning (*a.k.a.* Representation learning) has emerged as new area of ML research [43–45] that exploits multiple layers of information-processing in a hierarchical architecture for pattern classification and or representation learning (e.g., Feed-forward neural networks) [46]. The main advantage of deep learning is referred to the drastically increased chip processing abilities, the lowered cost of computing hardware, and the recent advances in ML.

Deep neural networks (DNNs) are multilayer networks with many hidden layers, whose weights are fully connected and often initialized or pretrained using stacked Restricted Boltzmann Machine (RBM) or Deep Belief Networks (DBMs) [46]. DBM is a pretraining unsupervised step that utilizes large amount of unlabeled training data for extracting structures and regularities in input features [47]. DBN not only uses a huge amount of unlabeled training data but also provides good initialization weights for DNN. Moreover, overfitting and underfitting problems can be tackled by using the pretraining step of DBN. DNN has shown great performance in recognition and classification tasks, including natural language processing, image classification, and traffic flow detection [48]. However, DNN has high computational cost and difficult to scale [49]. DSN addresses the scalability problem of DNN, simple classifiers are stacked on top of each other in order to construct more complex classifier [50][51].

New techniques used in Sections 3.3 and 3.4 could fit to the problems of DNN naturally. The decision function of DNN is as follows:

$$P_j = \frac{\exp(x_j)}{\sum_k \exp(x_k)},$$

where $P_j$ represents the class probability and $x_j$ and $x_k$ represent the total input to units *j* and *k* respectively. The cross entropy is defined as follows:

$$C = \sum_j d_j \log(p_j),$$

where $d_j$ represents the target probability for output unit *j*, and $P_j$ is the probability output for *j* after applying the activation function [52]. Now, the new semiparametric model's approximation is approximated as:

$$\frac{\exp(c_j)}{\sum_k \exp(c_k)} \approx \frac{\exp(x_j)}{\sum_k \exp(x_k)},$$

this approximation no longer extracts worst-case example *x* and is now able to reduce its complexity effectively. As in the local learning strategy, the model assumptions gets stronger than those of nonparametric models, but they are less restrictive than those of parametric model while reducing its computational complexity significantly.

### 3.6. Big data computing

Big data computing systems fall into two major categories, based on how data is analyzed with regards to time constraint [53]. First, batch processing of large volumes of on-disk data with no time constraints (e.g., MapReduce and GraphLab). Second, streaming processing of in-memory data in real-time or short period of time (e.g., Storm, SAMOA) [54][55]. In [54], Huang and Li argued that next-generation computing systems for big data analytics need innovative designs in both hardware and software that would provide a good match between big data algorithms and the underlying computing and storage resources.

There are several computing frameworks, e.g., Hadoop [56], SHadoop [57], ComMapReduce [58], Dryad [59], Piccolo [60], and IBM parallel machine learning toolbox, such systems have the capabilities to scale up machine learning. The combination of deep learning and parallel training implementation techniques provides potential ways to process Big Data [61]. Quoc V. Le *et al.* [62] consider the problem of building high-level, class-specific feature detectors from only unlabeled data. Experimental results reveal that it is possible to train a face detector without having to label images as containing a face or not.

K. Zhang and X. Chen [63] presented a distributed learning paradigm for the RBMs and the backpropagation algorithm using MapReduce. The DBNs are trained in a distributed way by stacking a series of distributed RBMs for pre-training and a distributed backpropagation for fine-tuning. Experimental results demonstrate that the distributed RBMs and DBNs are amenable to large-scale data with a good performance in terms of accuracy and efficiency.

## 4. Concluding Remarks

In this review, we provided an overview of the current state of research in sustainable data modeling. In particular, we discussed its theoretical and experimental aspects in large-scale data-intensive fields, relating to: (1) model energy efficiency, including computational requirements in learning, and possible approaches, and (2) data-intensive areas' structure and design, including the relation between data models and characteristics, With the surge in e-science data, sustainable data modeling has been shown to offer a way forward due to its ease in handling large quantities of data. It is also envisaged that such data-modeling revolution can be readily extended to various areas in e-science. These newly designed sustainable data models will not only be able to cope with the emerging large-scale data paradigm, but also provide a means in maximizing its return for the various e-science areas.

ACKNOWLEDGMENT

The authors are grateful to Dr. Jason W.P Ng at EBTIC, for his invaluable discussions and feedback, and special thanks to the British Telecom (BT) in London for their constructive criticism on this work.